\documentclass[runningheads]{llncs}
\usepackage[T1]{fontenc}
\usepackage{float}
\usepackage{amsmath}
\usepackage{mathtools}
\usepackage{amssymb}
\usepackage{graphicx}
\begin{document}
\title{Deep3DSketch+: Rapid 3D Modeling from Single Free-hand Sketches}
%
%
\author{Tianrun Chen\inst{1}\and Chenglong Fu\inst{2}\and Ying Zang*\inst{2} \and Lanyun Zhu\inst{3}\and Jia Zhang\inst{4}\and Papa Mao\inst{5}\and Lingyun Sun\inst{1}}
\authorrunning{T. Chen et al.}
%
\institute{College of Computer Science and Technology, Zhejiang University, Hangzhou 310027, China.\and
School of Information Engineering, Huzhou University, Huzhou 313000, China.\and
Information Systems Technology
and Design Pillar, Singapore University of Technology and Design, Singapore 487372, Singapore. \and Yangzhou Polytechnic College, Yangzhou 225009, China. \and Mafu Laboratory, Moxin (Huzhou) Tech. Co., LTD, Huzhou 313000, China.\\
\email{* 02750@zjhu.edu.cn}}
\maketitle              
\begin{abstract}
The rapid development of AR/VR brings tremendous demands for 3D content. While the widely-used Computer-Aided Design (CAD) method requires a time-consuming and labor-intensive modeling process, sketch-based 3D modeling offers a potential solution as a natural form of computer-human interaction. However, the sparsity and ambiguity of sketches make it challenging to generate high-fidelity content reflecting creators' ideas. Precise drawing from multiple views or strategic step-by-step drawings is often required to tackle the challenge but is not friendly to novice users. In this work, we introduce a novel end-to-end approach, Deep3DSketch+, which performs 3D modeling using only a single free-hand sketch without inputting multiple sketches or view information. Specifically, we introduce a lightweight generation network for efficient inference in real-time and a structural-aware adversarial training approach with a Stroke Enhancement Module (SEM) to capture the structural information to facilitate learning of the realistic and fine-detailed shape structures for high-fidelity performance. Extensive experiments demonstrated the effectiveness of our approach with the state-of-the-art (SOTA) performance on both synthetic and real datasets.

\keywords{Sketch\and 3D reconstruction\and 3D modeling.}
\end{abstract}
\section{Introduction}
\label{sec:intro}
The era has witnessed tremendous demands for 3D content \cite{wang2020vr}, especially with the rapid development of AR/VR and portable displays. Conventionally, 3D content is created through manual designs using Computer-Aided Design (CAD) methods. Designing numerous models by hand is not only labor-intensive and time-consuming, but also comes with high demands for the skill set of designers. Specifically, existing CAD-based methods require creators to master sophisticated software commands (\textit{commands knowledge}) and further be able to parse a shape into sequential commands (\textit{strategic knowledge}), which restricts its application in expert users \cite{bhavnani1999strategic,chester2007teaching}.

The restrictions of CAD methods call for the urgent need for alternative ways to support novice users to have access to 3D modeling. Among many alternatives, sketch-based 3D modeling has been recognized as a potential solution in recent years -- sketches play an important role in professional designing and our daily life, as it is one of the most natural ways we humans express ideas. Despite many works have utilized sketch to produce 3D models, the majority of existing efforts either require accurate line-drawings from multiple viewpoints or apply step-by-step workflow that requires \textit{strategic knowledge} \cite{li2020sketch2cad,cohen1999interface,deng2020interactive}, which is not user-friendly for the masses. Other work proposed retrieval-based approaches from existing models, which lack customizability. 

To mitigate the research gap, we aim to propose an effective method that uses only one single sketch as the input and generates a complete and high-fidelity 3D model. By fully leveraging the information from input human sketches, the designed approach should offer an intuitive and rapid 3D modeling solution to generate high-quality and reasonable 3D models that accurately reflects the creators' ideas. 

However, it is a non-trivial task to obtain high-quality 3D models from a single sketch. A significant domain gap exists between sketches and 3D models, and the sparsity and ambiguity of sketches bring extra obstacles. As in \cite{zhang2021sketch2model,guillard2021sketch2mesh}, deploying widely-used auto-encoder as the backbone of the network can only obtain coarse prediction, thus Guillard et al. \cite{guillard2021sketch2mesh} use post-processing optimization to obtain fine-grained mesh, which is a time-consuming procedure. It remains a challenge to have rapid 3D modeling from single sketches with high fidelity.

Facing the challenge, we hereby propose Deep3DSketch+, an end-to-end neural network with a lightweight generation network and a structural-aware adversarial training approach. Our method comes with a shape discriminator with input from the predicted mesh and ground truth models to facilitate the learning of generating reasonable 3D models. A Stroke Enhancement Module (SEM) is also introduced to boost the capability for structural feature extraction of the network, which is the key information in sketches and the corresponding silhouettes. Extensive experiments were conducted and demonstrated the effectiveness of our approach. We have reported state-of-the-art (SOTA) performance in both synthetic and real datasets.

\section{Related Works}
\subsection{Sketch-based 3D Modeling}
Sketch-based 3D modeling is a research topic that researchers have studied for decades. \cite{bonnici2019sketch,olsen2009sketch} review the existing sketch-based 3D modeling approaches. Existing sketch-based 3D modeling falls into two categories: end-to-end approach and interactive approach. The interactive approaches require sequential step decomposition or specific drawing gestures or annotations \cite{igarashi2006teddy,li2020sketch2cad,cohen1999interface,shtof2013geosemantic,jorge2003gides++,gingold2009structured,deng2020interactive}, in which users need to have strategic knowledge to perform the 3D modeling process. For end-to-end approaches, works that use template primitives or retrieval-based approaches \cite{chen2003visual,wang2015sketch,sangkloy2016sketchy,huang2016shape} can produce some decent results but lack customizability. Some very recent works \cite{zhang2021sketch2model,guillard2021sketch2mesh,wang20203d} directly reconstruct the 3D model using deep learning and recognized the problem as a single-view 3D reconstruction task. However, sketch-based modeling and conventional monocular 3D reconstruction have substantial differences -- the sparse and abstract nature of sketches and lack of textures calls for extra clues to produce high-quality 3D shapes, which are in this work we aim to solve.

\subsection{Single-view 3D Reconstruction}
Single-view 3D reconstruction is a long-standing and challenging task. Recent advances in large-scale datasets like ShapeNet~\cite{chang2015shapenet} facilitate rapid development in the field, making possible data-driven approaches. Among the data-driven methods, some ~\cite{chen2019learning,park2019deepsdf,mescheder2019occupancy,fan2017point,pontes2018image2mesh,girdhar2016learning,wang2018pixel2mesh} use category-level information to infer 3D representation from a single image. Others~\cite{liu2019soft,liu2019learning,loper2014opendr,kato2018neural,insafutdinov2018unsupervised,li2018differentiable} obtain 3D models directly from 2D images, in which the emergence of differentiable rendering techniques played a critical role. There are also recent advances~\cite{lin2020sdf,ke2017srn,mildenhall2020nerf,xu2019disn,yu2021pixelnerf} use unsupervised methods for implicit function representations by differentiable rendering. Many geometric processing approaches can enhance the performance. \cite{dou2020top,dou2022coverage,dou2022tore,xu2023globally,lin2023patch,xu2022rfeps,zhang2023surface} Whereas existing methods concentrate on learning 3D geometry from 2D images, we aim to obtain 3D meshes from 2D sketches -- a more abstract and sparse form than real-world colored images. Generating high-quality 3D shapes from such an abstract form of image representation is still a challenge that needs to be solved. 

\section{Method}

\begin{figure}[t]
\includegraphics[width=\linewidth]{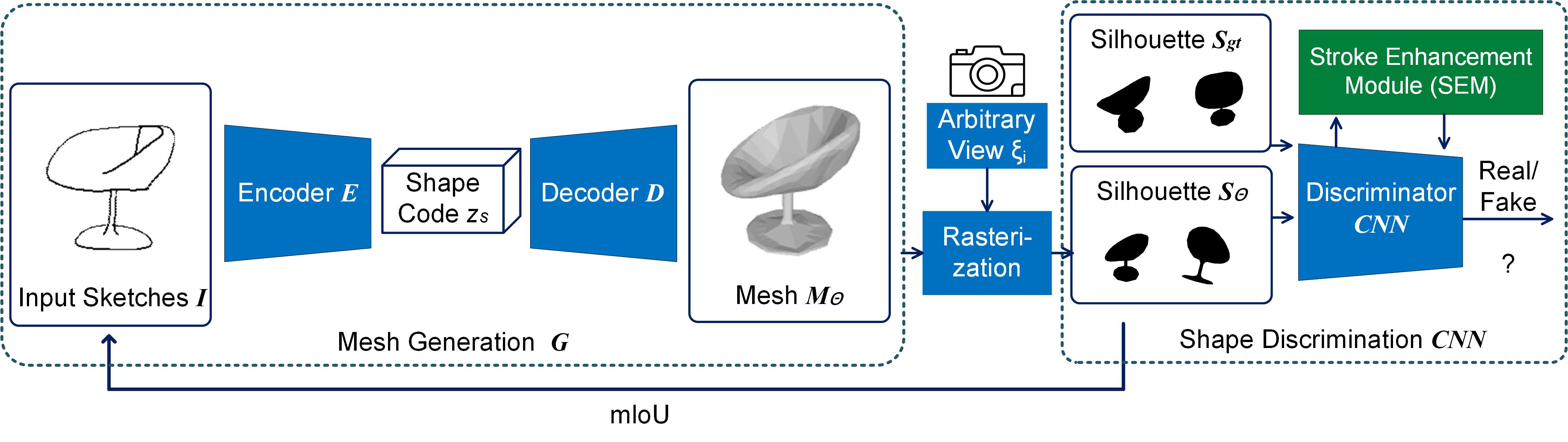}
\caption{{The overall structure of Deep3DSketch+.}} \label{fig:0}
\end{figure}

\subsection{Preliminary}

A single binary sketch ${I\in \left\{0,1\right\}^{W\times H}}$ is used as the input of 3D modeling. We let ${I \left [ i,j \right ] = 0 }$ 
if marked by the stroke, and ${I\left [ i,j \right ] = 1 }$ otherwise. 
The network $G$ is designed to obtain a mesh ${M_\Theta =(V_\Theta,  F_\Theta)}$, 
in which ${V_\Theta}$ and ${F_\Theta}$ represents the mesh vertices and facets, and the silhouette 
${S_\Theta}$ of ${M_\Theta}$ matches with the input sketch $I$. 

\subsection{View-aware and Structure-aware 3D Modeling.}

The overall structure of our method, Deep3DSketch+, is illustrated in Figure \ref{fig:0}. 
The backbone of the network $G$ is an encoder-decoder structure. As sketches are a sparse and ambiguous form of input, an encoder $E$ first transforms the input sketch into a latent shape code $z_s$, which summarizes the sketch on a coarse level with the involvement of the semantic category and the conceptual shape. A decoder $D$ consisting of cascaded upsampling blocks is then used to calculate the vertex offsets of a template mesh and deforms it to get the output mesh $M_\Theta = D(z_s)$ with fine details by gradually inferring the 3D shape information with increased spatial resolution. Next, the generated mesh $M_\Theta$ is rendered with a differentiable renderer and generates a silhouette $S_\Theta$. The network is end-to-end trained with the supervision of rendered silhouettes through approximating gradients of the differentiable renderer. 

However, due to the sparse nature of sketches and the only supervision of the single-view silhouette constraint, the encoder-decoder structured generator \textit{\textbf{G}} cannot effectively obtain high-quality 3D shapes. Extra clues must be used to attend to the fine-grained, and realistic objects' structures \cite{zhang2021sketch2model,guillard2021sketch2mesh}. Therefore, \cite{guillard2021sketch2mesh} introduces a two-stage post-refinement scheme through optimization, which first obtains a coarse shape and further optimizes the shape to fit the silhouette. However, such an approach is time-consuming and cannot meet the requirement of real-time interactive modeling. On the contrary, we aim to end-to-end learn a rapid mesh generation while also being capable of producing high-fidelity results. We introduce a shape discriminator and a stroke enhancement module to make it possible.

\noindent \textbf{Shape discriminator and Multi-view Sampling.}
 We aim to address the challenge by introducing a shape discriminator \textit{CNN}, which introduces the 3D shapes from real datasets during training to force the mesh generator \textit{\textbf{G}} to produce realistic shapes, while keeping the generation process efficient during inference. Specifically, the discriminator CNN is inputted with the generated silhouette from the predicted mesh and rendered silhouette from the manually-designed mesh. 

Moreover, we argue that a single silhouette cannot fully represent the information of the mesh because, unlike the 2D image translation task, the generated mesh $M_\Theta$ is a 3D shape that can be viewed in various views. The silhouette constraints ensure the generated model matches the viewpoint of the particular sketch but cannot guarantee the model is realistic and reasonable across views. Therefore, we propose to randomly sample $N$ camera poses $\xi_{1...N}$ from camera pose distribution $ p_{\xi}$. Findings in the realm of shape-from-silhouette have demonstrated that multi-view silhouettes contain valuable geometric information about the 3D object~\cite{gadelha2019shape,hu2018structure}. We use a differentiable rendering module to render the silhouettes $S_{1...N}$ from the mesh $M$ and render the silhouettes $S_r\left \{1...N \right \} $ from the mesh $M_r$. The differentiable rendering equation $R$ is shown in \cite{liu2019soft}. 

By inputting the $S_r\left \{1...N \right \} $ to the discriminator for the predicted meshes and the real meshes, the network is aware of the geometric structure of the objects in cross-view silhouettes, ensuring the generated mesh is reasonable and high-fidelity in detail.  

\begin{figure}[t]
\includegraphics[width=\linewidth]{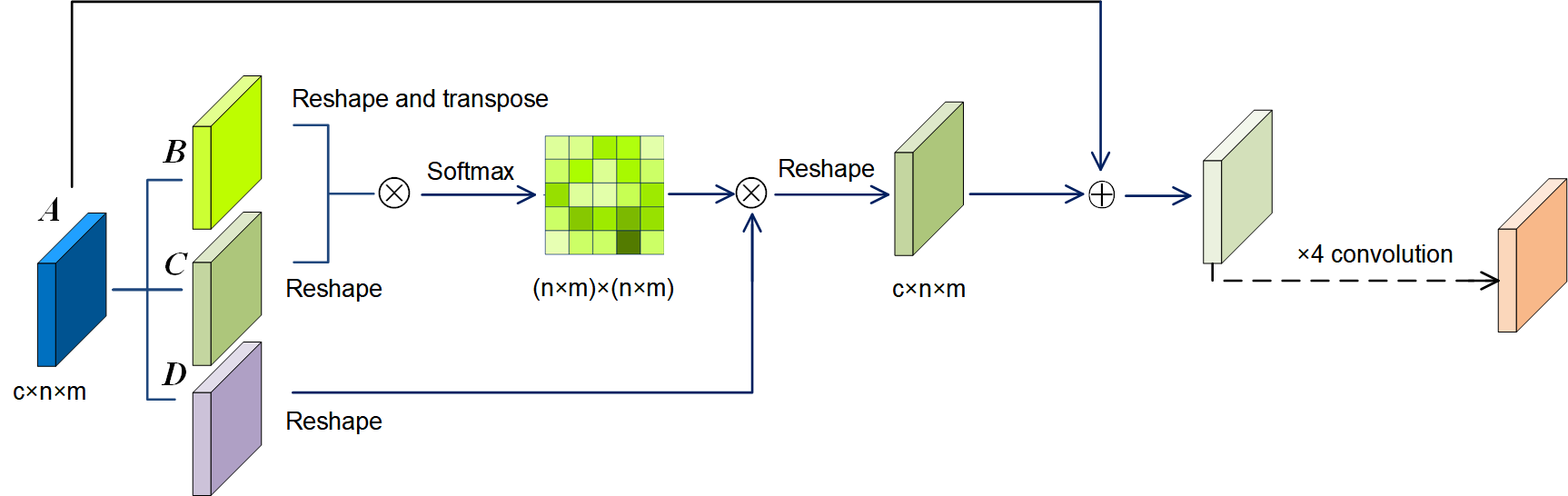}
\caption{{The Details of Stroke Enhancement Module (SEM). $\otimes$ denotes element-wise multiplication, $\oplus$ demotes element-wise add operation.}} \label{fig:1}
\end{figure}

\noindent\textbf{Stroke Enhancement Module. } 
Sketch-based 3D modeling differs from conventional monocular 3D reconstruction tasks, in which the input image has rich textures and versatile features for predicting depth information. But in our sketch-based modeling task, the input sketch and projected silhouettes are in a single color and thus cannot effectively obtain depth prediction results. Alternatively, we propose to fully utilize the monocolored information for feature extraction by introducing a stroke enhancement module (SEM), as shown in Figure \ref{fig:1}. The SEM consists of a position-aware attention module as in \cite{fu2019dual} that encodes a wide range of contextual information into local features to learn the spatial interdependencies of features \cite{chen2020supervised} and a post-process module that is designed to manipulate the feature from position-aware attention with a series of convolutions in order to smoothly add them to the original feature before attention in an element-wise manner. Such a strategy can boost the learning of features in the targeted positions, especially on the boundary. 
Specifically, the local feature from the silhouette $A \in \mathbb{R}^{C \times N \times M}$ is fed into a convolutional layer to form two local features
$B, C \in \mathbb{R}^{C\times W}$ where $W=M\times N$ equals the number of pixels, and another convolutional layer is used to form the feature map $D \in \mathbb{R}^{C\times N\times M}$. Matrix multiplication is performed between the transpose of $C$ and $B$, followed by a softmax layer to generate the attention map $S \in \mathbb{R}^{W\times W}$, thus enhancing the capability of the utilization of key structural information represented by the silhouette.

\begin{equation}
s_{ij} = \frac{\exp\left(B_i * C_j\right)}{\sum_{i=1}^{W}\exp\left(B_i * C_j\right)},
\end{equation}

The attention map is used to produce the output $F$ through a weighted sum of the original feature and the features across all positions, 
\begin{equation}
    F_{j} = {\lambda\sum_{i=1}^{W}\left(s_j · D_j\right)+A_j}
\end{equation}

\subsection{Loss Function}
The loss functions are carefully designed with three components to train the network: a multi-scale mIoU loss $\mathcal{L}_{sp}$, flatten loss and laplacian smooth loss $\mathcal{L}_{r}$, and a structure-aware GAN loss $\mathcal{L}_{sd}$.
The multi-scale mIoU loss $\mathcal{L}_{sp}$ measures the similarity between rendered silhouettes and ground truth silhouettes. Aiming at improving computational efficiency, we progressively increase the resolutions of silhouettes, which is represented as 
\begin{align}
\mathcal{L}_{s p}=\sum_{i=1}^{N} \lambda_{s_{i}} \mathcal{L}_{iou}^{i}
\label{lsp}
\end{align}

 $\mathcal{L}_{iou}$ is defined as:
\begin{align}
\mathcal{L}_{i o u}\left(S_{1}, S_{2}\right)=1-\frac{\left\|S_{1} \otimes S_{2}\right\|_{1}}{\left\|S_{1} \oplus S_{2}-S_{1} \otimes S_{2}\right\|_{1}}
\end{align}
where $S_1$ and $S_2$ is the rendered silhouette. 

We also proposed to use flatten loss and Laplacian smooth loss to make meshes more realistic with higher visual quality, represented by $\mathcal{L}_{r}$, as shown in~\cite{zhang2021sketch2model,kato2018neural,liu2019soft}.

For our structure-aware GAN loss  $\mathcal{L}_{sd}$, non-saturating GAN loss \cite{mescheder2018training} is used.
\begin{align}
\begin{split}
\mathcal{L}_{sd} &=\mathbf{E}_{\mathbf{z_v} \sim p_{z_v}, \xi \sim p_{\xi}}\left[f\left(CNN_{\theta_{D}}\left(R(M, \xi)\right)\right)\right] \\
&+\mathbf{E}_{\mathbf{z_{vr}} \sim p_{z_{vr}}, \xi \sim p_{\xi}}\left[f\left(-CNN_{\theta_{D}}(R(M_r, \xi))\right)\right] \label{gan}
\end{split}\\ 
& \mathit{{ where  }}  f(u)=-\log (1+\exp (-u))
\end{align}

The overall loss function $Loss$ is calculated as the weighted sum of the three components:
\begin{align}
Loss =  \mathcal{L}_{sp} + \mathcal{L}_{r} + \lambda_{sd} \mathcal{L}_{sd}
\label{loss}
\end{align}
\section{Experiments}

\subsection{Datasets}
Public available dataset for sketches and the corresponding 3D models is rare. Following \cite{zhang2021sketch2model}, we take an alternative solution by using the synthetic data \textit{\textbf{ShapeNet-synthetic }}for training, and apply the trained network to real-world data \textit{\textbf{ShapeNet-sketch}} for performance evaluation. 
The synthetic data is obtained by collecting an edge map extracted by a canny edge detector of rendered images provided by Kar et al. \cite{kar2017learning}. 13 categories of 3D objects from ShapeNet are used. The ShapeNet-Sketch is collected by real-human. Volunteers with different drawing skills draw objects based on the images of 3D objects from \cite{kar2017learning}. A total number of 1300 sketches and their corresponding 3D shapes are included in the dataset.  
\begin{table}[!htb]
\caption{The quantitative evaluation of ShapeNet-Synthetic dataset.}
\begin{center}{
\begin{tabular}{|c|c|c|c|c|c|c|c|}
\hline
\multicolumn{8}{|c|}{Shapenet-synthetic (Voxel Iou $\uparrow)$} \\
\hline
 & car & sofa & airplane & bench & display & chair & table \\
\hline
Retrieval & 0.667 & 0.483 & 0.513 & 0.380  & 0.385 & 0.346 & 0.311 \\
\hline
Auto-encoder & 0.769 & 0.613 & 0.576 & 0.467 & 0.541 & 0.496 & \textbf{0.512} \\
\hline
Sketch2Model  & 0.751 & 0.622 & 0.624 & 0.481 & \textbf{0.604} & 0.522 & 0.478 \\
\hline
\textbf{Ours} & \textbf{0.782} & \textbf{0.640} & \textbf{0.632} & \textbf{0.510} & 0.588 & \textbf{0.525} & 0.510 \\
\hline
 & telephone & cabinet & loudspeaker & watercraft & lamp & rifile & mean \\
\hline
Retrieval & 0.622 & 0.518 & 0.468 & 0.422 & 0.325 & 0.475 & 0.455 \\
\hline
Auto-encoder & 0.706 & 0.663 & 0.629 & 0.556 & 0.431 & 0.605 & 0.582 \\
\hline
Sketch2Model & 0.719 & \textbf{0.701} & \textbf{0.641} & \textbf{0.586} & \textbf{0.472} & 0.612 & 0.601 \\
\hline
\textbf{Ours} & \textbf{0.757} & 0.699 & 0.630 & 0.583 & 0.466 & \textbf{0.632} & \textbf{0.611} \\ 
\hline
\end{tabular}}
\end{center}

\label{table:table1}

\end{table}

\subsection{Implementation details}
We use ResNet-18 \cite{he2016deep} for the encoder $E$ for image feature extraction. SoftRas \cite{liu2019soft} is used for rendering silhouettes. Each 3D object is placed with 0 in evaluation and 0 in azimuth angle in the canonical view, with a fixed distance from the camera. The ground-truth viewpoint is used for rendering. Adam optimizer with the initial learning rate of 1e-4 and multiplied by 0.3 for every 800 epochs. Betas are equal to 0.9 and 0.999. The total training epochs are equal to 2000. The model is trained individually with each class of the dataset. $\lambda_{sd}$ in Equation. \ref{loss} equal to 0.1. 

\subsection{Results}
\begin{figure*}[h]
\centering
\includegraphics[width=\linewidth]{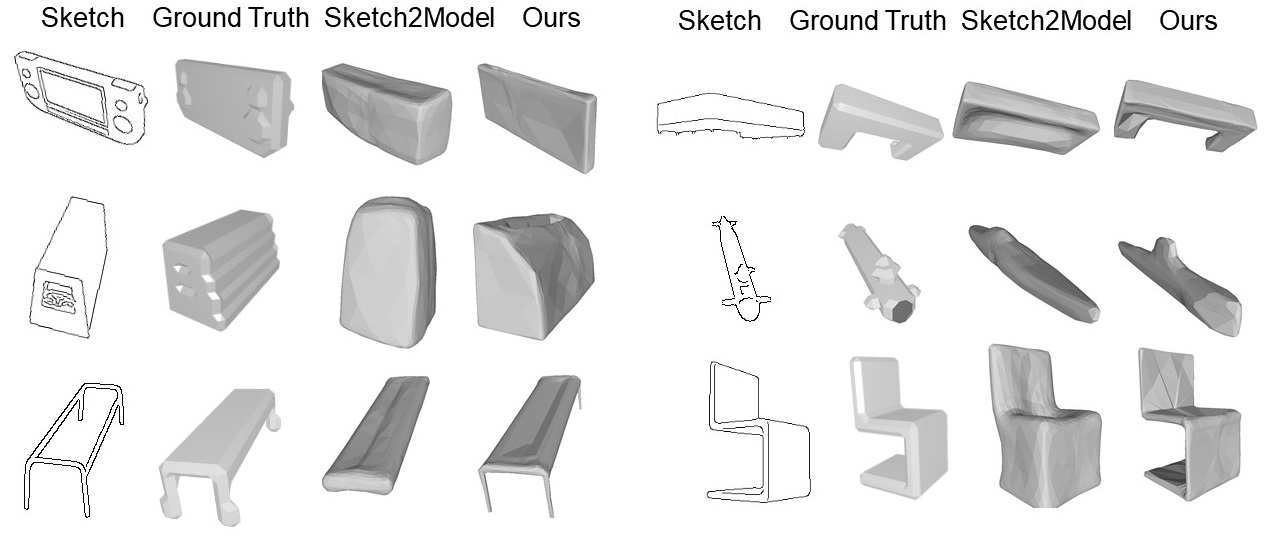}
\caption{{\textbf{Qualitative evaluation with existing state-of-the-art.} The visualization of 3D models generated demonstrated that our approach is capable of obtaining higher fidelity of 3D structures.}}
\label{fig:2}
\end{figure*}
\noindent \textbf{The ShapeNet-Synthetic Dataset.} 

Following \cite{zhang2021sketch2model}, we compare our method with the model retrieval approach with features from a pre-trained sketch classification network, and the \cite{zhang2021sketch2model} as the existing state-of-the-art (SOTA) model. We first evaluate the model performance on the\textit{ ShapeNet-Synthetic} dataset, which has the accurate ground truth 3D model for training and evaluation. Commonly-used 3D reconstruction metrics -- voxel IoU is used to measure the fidelity of the generated mesh, as shown in Table \ref{table:table1}. We also measured the Chamfer Distance, another widely used metric for mesh similarities, as shown in the Supplementary Material. The quantitative evaluation shows the effectiveness of our approach, which achieves state-of-the-art (SOTA) performance. We also conducted a quantitative evaluation of our method compared with existing state-of-the-art, which further demonstrated the effectiveness of our approach to producing models with higher quality and fidelity in structure, as shown in Figure \ref{fig:2}.

\noindent \textbf{The ShapeNet-Sketch Dataset.} 

After training in the synthetic data, we further evaluate the performance of real-world human drawings, which is more challenging due to the creators' varied drawing skills and styles. A domain gap also exists in the synthetic and real data when we train the model on ShapeNet-Synthetic Dataset and use ShapeNet-Sketch Dataset for evaluation. In such settings, a powerful and robust feature extractor with structural-awareness is more critical. In experiments, our model generalizes well in real data. As shown in Table \ref{table:table2}, our model outperforms the existing state-of-the-art in most categories, demonstrating the effectiveness of our approach. It is worth noting that the domain adaptation technique could be a potential booster for the network's performance in real datasets with the domain gap in presence, which could be explored in future research.

\noindent \textbf{Evaluating Runtime for 3D modeling.}

As previously mentioned, we aim to make the network efficient for rapid 3D modeling. After the network was well-trained, We evaluated the neural network on a PC equipped with a consumer-level graphics card (NVIDIA GeForce RTX 3090). Our method achieves the generation speed of 90 FPS, which is a 38.9$\%$ of speed gain compared to Sketch2Model (0.018s) \cite{zhang2021sketch2model}.  We also tested the performance solely on CPU (Intel Xeon Gold 5218) and reported an 11.4$\%$ of speed gain compared to Sketch2Model (0.070s)  \cite{zhang2021sketch2model}, with the rate of 16FPS, which is sufficient to be applied for smooth computer-human interaction. 
\begin{table}[H]
\setlength{\tabcolsep}{7mm}
\caption{Average Runtime for Generatin a Single 3D Model.}
\begin{center}
\scalebox{0.9}{
\begin{tabular}{ c || c | c || c}
\hline
 \textbf{Inference by GPU} & 0.011 s & \textbf{Inference by CPU} & 0.062 s \\
\hline
\end{tabular}}
\end{center}
\label{table:table3}

\end{table}


\begin{table}[!htb]
\caption{The quantitative evaluation of ShapeNet-Sketch dataset.}
\begin{center}{
\begin{tabular}{|c|c|c|c|c|c|c|c|}
\hline
\multicolumn{8}{|c|}{Shapenet-sketch (Voxel Iou $\uparrow$)} \\
\hline
 & car & sofa & airplane & bench & display & chair & table \\
\hline
Retrieval & 0.626 & 0.431 & 0.411 & 0.219 & 0.338 & 0.238 & 0.232 \\
\hline
Auto-encoder & 0.648 & \textbf{0.534} & 0.469 & 0.347 & 0.472 & 0.361 & 0.359 \\
\hline
Sketch2Model  & 0.659 & \textbf{0.534} & 0.487 & 0.366 &  \textbf{0.479} &  \textbf{0.393} & 0.357\\
\hline
\textbf{Ours} &  \textbf{0.675} &  \textbf{0.534} &  \textbf{0.490} & \textbf{0.368} & 0.463 & 0.382 &  \textbf{0.370} \\
\hline
 & telephone & cabinet & loudspeaker & watercraft & lamp & rifile & mean \\
\hline
Retrieval & 0.536 & 0.431 & 0.365 & 0.369 & 0.223 & 0.413 & 0.370 \\
\hline
Auto-encoder & 0.537 & 0.534 &  0.533 & 0.456 & 0.328 & 0.541 & 0.372 \\
\hline
Sketch2Model & 0.554 & \textbf{0.568} & \textbf{0.544} & 0.450 & 0.338 & 0.534 & \textbf{0.483} \\
\hline
\textbf{Ours} & \textbf{0.576} & 0.553 & 0.514 & \textbf{0.467} &  \textbf{0.347} &  \textbf{0.543} &  \textbf{0.483} \\
\hline
\end{tabular}}
\end{center}
\label{table:table2}

\end{table}





\subsection{Ablation Study}

\begin{figure*}[h]
\centering
\resizebox{.6\textwidth}{!}{
\includegraphics[width=\linewidth]{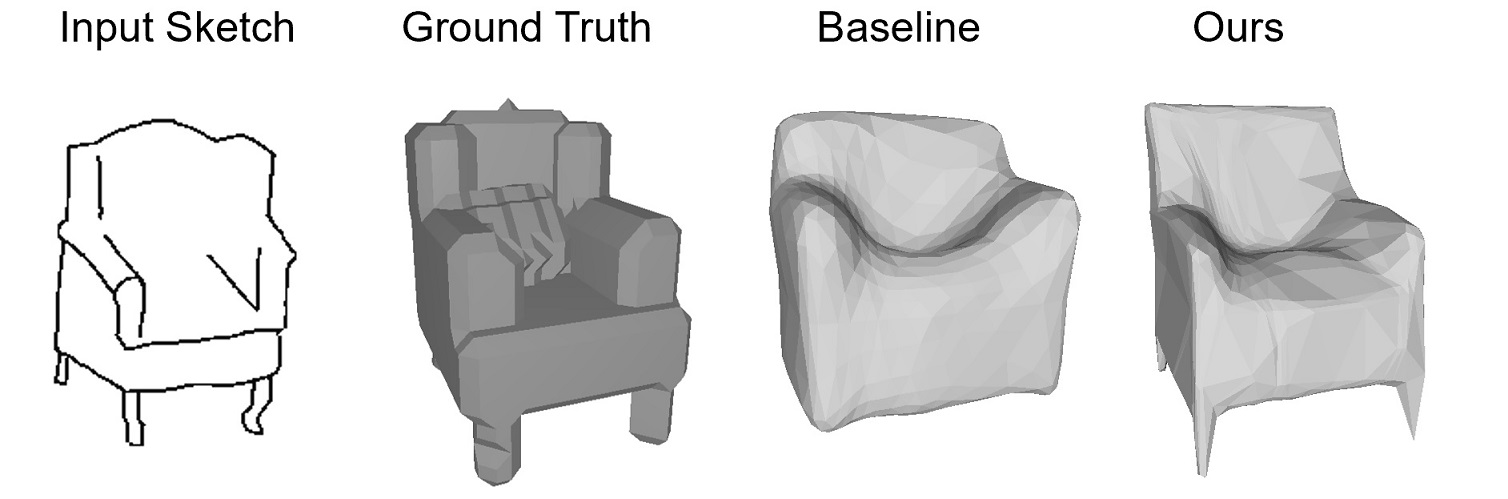}}
\caption{{\textbf{Ablation Study.} Our method generates more fine-grained structures compared to the baseline method.}}
\label{fig:4}
\vspace{0.1cm}
\end{figure*}

To verify the effectiveness of our proposed method, we conducted the ablation study as shown in Table \ref{table:table3}. We demonstrated that our method with Shape Discriminator (SD) and Stroke Enhancement Module (SEM) contributed to the performance gain to produce models with higher-fidelity, as shown in Figure. \ref{fig:4}, compared to w/o SD or SEM (baseline method).

\begin{table}[!htb]
\caption{Ablation Study.}
\begin{center}
\scalebox{0.9}{
\begin{tabular}{c c || c c c c c c c }
\hline
SD & SEM & car & sofa & airplane & bench & display & chair & table \\
\hline
\hline
 & & 0.767  & 0.630  & 0.633 & 0.503 & 0.586 & 0.524  & 0.493  \\
$\surd$ & & 0.778  & 0.632 & \textbf{0.637} & 0.503 & \textbf{0.588} & 0.523  & 0.485  \\
$\surd$ & $\surd$ & \textbf{0.782}  & \textbf{0.640}  & 0.632 & \textbf{0.510}  & \textbf{0.588}  & \textbf{0.525}  & \textbf{0.510}  \\
\hline
SD & SEM & telephone & cabinet & loudspeaker & watercraft & lamp & rifile & mean \\
\hline
\hline
 & & 0.742 & 0.690 & 0.555 & 0.563 & 0.458 & 0.613 & 0.598 \\
$\surd$ & & 0.749  & 0.688  & 0.617 & 0.567 & 0.454 & 0.612  & 0.602  \\
$\surd$ & $\surd$ & \textbf{0.757}  & \textbf{0.699} & \textbf{0.630}  & \textbf{0.583}  & \textbf{0.466} & \textbf{0.624}  & \textbf{0.611}  \\
\hline
\end{tabular}}
\end{center}
\label{table:table3}

\end{table}

\begin{figure*}[h]
\centering
\includegraphics[width=0.8\linewidth]{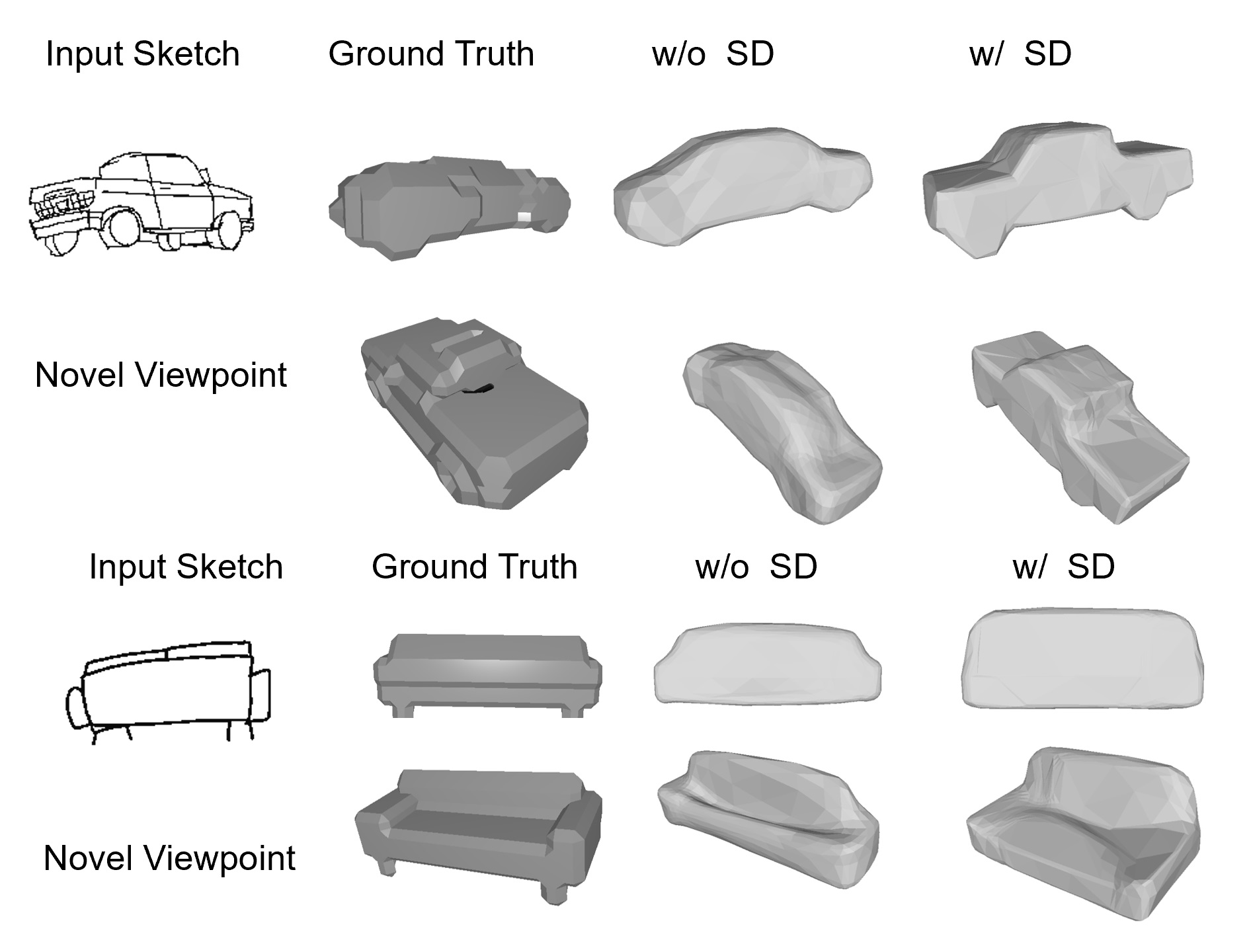}
\caption{{\textbf{The effectiveness of SD and random-viewpoint sampling.} As shown in the example, the neural network generates more fine-grained structures compared to the baseline method.}}
\label{fig:5}
\end{figure*}

Moreover, We argue that the random viewpoint sampling combined with the shape discriminator (SD) with real shapes as inputs allows the neural network "to see" real shapes from multiple angles, thus being capable of predicting reasonable structural information that is not even in presence in the sketch (which might be not represented due to the viewpoint constraints). In Figure. \ref{fig:5}, We show several examples. It can be observed that the region of the sitting pad on the sofa is reconstructed, although the input human sketch is only viewed backward. The flat plane at the back of the car is reconstructed, although the input human sketch is only viewed near the front, thanks to the introduction of SD and random viewpoint sampling.

\section{Conclusion}
In this paper, we propose Deep3DSketch+, which takes a single sketch and produces a high-fidelity 3D model. We introduce a shape discriminator with random-pose sampling to allow the network to generate reasonable 3D shapes and a stroke enhancement model to fully exploit the mono-color silhouette information for high-fidelity 3D reconstruction. The proposed method is efficient and effective, and it is demonstrated by our extensive experiments --  we have reported state-of-the-art (SOTA) performance on both real and synthetic data. We believe that our proposed easy-to-use and intuitive sketch-based modeling method have great potential to revolutionize future 3D modeling pipeline. 

\bibliographystyle{unsrt}
\bibliography{egbib}

\end{document}


\title{Supplementary Material for Deep3DSketch+: Rapid 3D Modeling from Single Free-hand Sketches}
%
%
\author{Anonymous submission}
\authorrunning{MMM submission 103}
%
\institute{
\email{}\\
\email{}}
\maketitle              
\section{The Evaluation of Chamfer Distance}
\begin{table}[!htb]
\caption{The evaluation of Chamfer Distance. Our method outperformed existing State-of-the-art method in most categories}
\begin{center}{
\begin{tabular}{|c|c|c|c|c|c|c|c|}
\hline
\multicolumn{8}{|c|}{Shapenet-synthetic (Chamfer Distance ( CD-l_2 × 1000 $\downarrow)$} \\
\hline
 & car & sofa & airplane & bench & display & chair & table \\
\hline

Sketch2Model (GT Pos)  & 6.323 & 24.14 & \textbf{5.849} & 11.71 & \textbf{9.663} & 18.95 & 18.99 \\
Sketch2Model (Pred Pos) & 6.667 & 22.88 & 6.204 & 12.20 & 15.37 & 19.23 & 19.43 \\
\hline
Ours & \textbf{5.424} & \textbf{22.65} &5.884 & \textbf{8.960} & 10.10	& \textbf{17.03} & \textbf{15.69} \\

\hline

 & telephone & cabinet & loudspeaker & watercraft & lamp & rifile & mean \\
\hline
Sketch2Model (GT Pos) & \textbf{6.441} & 11.04 & 14.86 & 12.81 & 51.49 & 6.285 & 15.27 \\
Sketch2Model (Pred Pos) & 9.489 & 14.44 & 16.64 & 14.36 & 51.86 & 6.551& 15.56 \\
\hline
Ours & 6.530 &\textbf{10.65} & \textbf{14.41} & \textbf{12.46 }& \textbf{46.78} & 6.434 &\textbf{ 14.07} \\ 

\hline
\end{tabular}}
\end{center}

\label{table:table4}

\end{table}

\section{More Implementation Details}
For each predicted model and GT model, we use N=2 for rendering. The silhouette with the corresponding ground truth is used for calculating the IoU loss. When rendering the silhouette of random viewpoint, we use a uniform camera distribution. We trained and evaluated our neural network on four NVIDIA GeForce RTX3090 graphics card.